\documentclass[twoside]{article}

\usepackage[accepted]{aistats2019}

\usepackage{amsmath, amsthm, amssymb}
\usepackage{apalike}
\usepackage{graphicx}
\usepackage{booktabs}
\usepackage{subcaption}
\usepackage[utf8]{inputenc}
\usepackage{tikz}
\usetikzlibrary{positioning}
\usetikzlibrary{calc}
\usepackage[outline]{contour}
\contourlength{1.3pt}
\usepackage{pgfplots}
\pgfplotsset{compat=1.14}

\usepackage[colorlinks, citecolor=blue]{hyperref}

\newcommand{\reals}{\mathbb{R}}
\newcommand{\Yk}{Y^{(k)}}
\newcommand{\Yki}{Y^{(k,1)}}
\newcommand{\Ykii}{Y^{(k,2)}}
\newcommand{\Zk}{Z^{(k)}}
\newcommand{\Zki}{Z^{(k,1)}}
\newcommand{\Zkii}{Z^{(k,2)}}
\newcommand{\tk}{\theta^{(k)}}
\newcommand{\vx}{\vec{x}}
\newcommand{\Fk}{F^{(k)}}
\newcommand{\fk}{f^{(k)}}
\newcommand{\tfk}{\tilde{f}^{(k)}}
\newcommand{\Lk}{L^{(k)}}

\newcommand{\evwrt}[2]{\mathbb{E}_{#1}\left[ #2 \right]}

\begin{document}

%

%

\twocolumn[

\aistatstitle{Deep Switch Networks for Generating Discrete Data and Language}

\aistatsauthor{Payam Delgosha \And Naveen Goela}

\aistatsaddress{ University of California, Berkeley \\ \texttt{pdelgosha@eecs.berkeley.edu} \And  Tanium, Data Science\\ \texttt{ngoela@alum.mit.edu} }

]

\begin{abstract} Multilayer switch networks are proposed as artificial generators of high-dimensional discrete data (e.g., binary vectors, categorical data, natural language, network log files, and discrete-valued time series). Unlike deconvolution networks which generate continuous-valued data and which consist of upsampling filters and reverse pooling layers, multilayer switch networks are composed of adaptive switches which model conditional distributions of discrete random variables. An interpretable, statistical framework is introduced for training these nonlinear networks based on a maximum-likelihood objective function. To learn network parameters, stochastic gradient descent is applied to the objective. This direct optimization is stable until convergence, and does not involve back-propagation over separate encoder and decoder networks, or adversarial training of dueling networks. While training remains tractable for moderately sized networks, Markov-chain Monte Carlo (MCMC) approximations of gradients are derived for deep networks which contain latent variables. The statistical framework is evaluated on synthetic data, high-dimensional binary data of handwritten digits, and web-crawled natural language data. Aspects of the model's framework such as interpretability, computational complexity, and generalization ability are discussed.
\end{abstract}

\section{Introduction}
\label{sec:introduction}

Several deep generative models have been proposed in the literature, including stacked Restricted Boltzmann Machines (RBMs), pixel convolutional neural networks (pixel CNNs), recurrent neural networks (RNNs), variational auto-encoders (VAEs), and generative adversarial networks (GANs). Many generative models produce continuous-valued data such as images. The present paper establishes a statistical model for generating high-dimensional discrete data. To learn discrete distributions, higher-order multivariate and long-range dependencies must be modeled effectively. A smooth interpolation between discrete samples is not guaranteed. For this reason, we propose an adaptive switch, instead of a filter, as the fundamental element of a generative model.

\subsection{Generative Modeling}

\subsubsubsection{\bf{RBMs:}} Discrete samples can be generated by latent-variable models such as RBMs using block Gibbs sampling. RBMs may be trained efficiently via contrastive divergence learning procedures~\cite{Hinton_Prod_of_Experts_Contrastive_2002}. Although training more sophisticated multilayer RBMs is feasible~\cite{pmlr-v5-salakhutdinov09a}, most approaches do not result in simple statistical models. The log-likelihood, even for single-layer RBMs, must be approximated due to the intractability of the partition function~\cite{Tieleman2008}.

\subsubsubsection{\bf{Pixel-CNNs, RNNs, Language Models:}} Several generative models of sequences involve optimizing convolutional filters via back-propagation of gradients. Pixel-CNNs are defined by filter transformations tailored for continuous-valued data~\cite{Conditional_PixelCNN_NIPS2016}. To harness existing training methods for discrete-valued data, language models represent each word in a vocabulary by a real-valued embedding  vector~\cite{JLMRBengio_2003,limits_of_NLP_2016,sparsely_gated_moe_NN_2017}. Embedding vectors may not be applicable to all discrete distributions. Many language models also view the generation of data conditioned on past variables as a problem of prediction or classification. In the absence of log-likelihoods, perplexity scores and performance indicators from downstream tasks serve as standards for assessment.

\subsubsubsection{\bf{Latent-Variable Models:}} VAEs and variational RNNs specify joint distributions over observed and latent variables~\cite{VAE_generative_model_14,NIPS2015_5653_Chung}. The posterior conditional density over latent variables is approximated using variational inference~\cite{Blei_Variational_2017}. Training VAEs involves the optimization of both encoder and decoder networks, approximations such as the evidence lower bound (ELBO), and the reparameterization trick for gradients which usually implies continuous latent variables. VAEs have been modified to include discrete latent variables and components~\cite{NIPS2017_vinyals,discrete_VAE_2017,NIPS2018_Vahdat}.




\subsubsubsection{\bf{GANs:}} Generative adversarial networks are optimized by adversarial training between a generator and a critic network~\cite{goodfellow2014generative}. More stable variants have been devised, including the Wasserstein GAN~\cite{salimans2016improved,arjovsky2017wasserstein}. Progressively-trained GANs have generated images with improved quality, resolution, and diversity~\cite{karras2018progressive}. Modifying GAN architectures for producing discrete data has yielded partial successes; e.g., the maximum-likelihood augmented GAN~\cite{MaliGAN_2017} and boundary-seeking GAN~\cite{devon2018boundary}. Several theoretical questions regarding the generalization and equilibrium of GANs have been studied, including whether the trained distribution is close to the target, and whether mode collapse can be prevented~\cite{pmlr-v70-arora17a,arora2018do}.


\subsection{Deep Switch Networks}

The present paper introduces nonlinear switching mechanisms as a means for learning discrete distributions. The switch model is simple, interpretable, and easily trained via a direct maximum-likelihood optimization. The training of deep switch networks which contain latent variables is feasible due to MCMC gradient approximations. We note that highway networks~\cite{HighwayNetworks_2015,RecurrentHighwayNetworks_2017}, and maxout networks~\cite{MaxoutNetworks_2013} contain network elements similar to adaptive switches. However, to the best of our knowledge, deep switch networks have not been studied previously for generative modeling of discrete data.


\begin{figure*}[t]
  \centering
  \vspace{0.21in}
  \begin{tikzpicture}
  \begin{scope}[scale=0.762,xshift=-5cm]
\definecolor{myblue}{rgb}{0.12156862745098,0.466666666666667,0.705882352941177}
\definecolor{myred}{rgb}{0.83921568627451,0.152941176470588,0.156862745098039}
\definecolor{myorange}{rgb}{1,0.498039215686275,0.0549019607843137}

\draw[white] (-5.6,-3.2) rectangle (5.6,3);

\node[circle, fill=myblue, inner sep = 3pt] (00) at (-1,-1) {};
\node[below=1mm of 00,scale=0.9] {$(0,0)$};
\node[ circle, fill=myblue, inner sep = 3pt] (11) at (1,1) {};
\node[above=1mm of 11,scale=0.9] {$(1,1)$};

\node[circle, fill=myred, inner sep = 3pt] (01) at (-1,1) {};
\node[above=1mm of 01,scale=0.9] {$(0,1)$};
\node[circle, fill=myred, inner sep = 3pt] (10) at (1,-1) {};
\node[below=1mm of 10,scale=0.9] {$(1,0)$};

\draw[->, very thick, black] (-2.5,-2.5) -- (2.5,-2.5) node[below=1mm,black] {$x_1$};
\draw[->, very thick, black] (-2.5,-2.5) -- (-2.5,2.5) node[left=1mm,black] {$x_2$};

\draw[very thick, black] (0,0) -- node[sloped, below, near end,black, scale=0.7] {$w_1 x_1 + w_2 x_2 + b = 0$} (20:3.5);
\draw[very thick, black] (0,0) -- (200:4) node[left, black] {$P=1/2$};
\draw[thick, ->, myred] ($(200:3) + (110:0.2)$) -- +(110:1) node[left=2mm] {$P>1/2$};
\draw[thick, ->, myblue] ($(200:3.5) + (290:0.2)$) -- +(290:1) node[left=2mm] {$P<1/2$};

\node at (0,-3.8) {$(a)$ \textit{Logistic Function} };
\end{scope}

  \begin{scope}[scale=0.762,xshift=5cm]
\definecolor{myblue}{rgb}{0.12156862745098,0.466666666666667,0.705882352941177}
\definecolor{myred}{rgb}{0.83921568627451,0.152941176470588,0.156862745098039}
\definecolor{myorange}{rgb}{1,0.498039215686275,0.0549019607843137}

\draw[white] (-5.6,-3.2) rectangle (5.6,3);

\node[circle, fill=myblue, inner sep = 3pt] (00) at (-1,-1) {};
\node[below=1mm of 00,scale=0.9] {$(0,0)$};
\node[ circle, fill=myblue, inner sep = 3pt] (11) at (1,1) {};
\node[above=1mm of 11,scale=0.9] {$(1,1)$};

\node[circle, fill=myred, inner sep = 3pt] (01) at (-1,1) {};
\node[above=1mm of 01,scale=0.9] {$(0,1)$};
\node[circle, fill=myred, inner sep = 3pt] (10) at (1,-1) {};
\node[below=1mm of 10,scale=0.9] {$(1,0)$};

\draw[->, very thick, black] (-2.5,-2.5) -- (2.5,-2.5) node[below=1mm,black] {$x_1$};
\draw[->, very thick, black] (-2.5,-2.5) -- (-2.5,2.5) node[left=1mm,black] {$x_2$};

\draw[very thick, black] (0,0) -- node[sloped, below, near end,black, scale=0.7] {$w^{(1)}_1 x_1 + w^{(1)}_2 x_2 + b^{(1)} = 0$} (20:3.5);
\draw[very thick, black] (0,0) -- (200:3.5);

\draw[very thick, black] (0,0) -- node[sloped, below, near end,black, scale=0.7] {\contour{white}{$w^{(2)}_1 x_1 + w^{(2)}_2 x_2 + b^{(2)} = 0$}} (160:3.5);
\draw[very thick, black] (0,0) --  (-20:3.5);

\draw[dashed, very thick, myorange] (0,-2.8) -- (0,3) node[below right=2mm,scale=0.7] {$\alpha_1 x_1 + \alpha_2 x_2 + \beta = 0$};

\node at (0,-3.8) {$(b)$ \textit{Nonlinear Adaptive Switch} };
\end{scope}

\end{tikzpicture}

\caption{ \textit{Elementary example:} Let $\{X_1, X_2\}$ denote two independent Bernoulli$(1/2)$ random variables, and let $X_3 = X_1 \oplus X_2$. Towards the goal of modeling the joint distribution of $\{X_1, X_2, X_3\}$, we model the conditional distribution $p_{X_3|X_1,X_2}(x_3| x_1, x_2)$. $(a)$ If the conditional is modeled by a single logistic function, the corresponding linear separator partitions $\reals^2$ into two half-planes. Accurate modeling requires distinguishing the set of red points from the set of blue points, which is impossible with a linear separator (Section~\ref{subsec:limitations_sigmoid}). $(b)$ An adaptive switch is optimal for generative modeling. For this specific example, the adaptive switch itself is modeled as a linear separator, $\alpha_1 x_1 + \alpha_2 x_2 + \beta = 0$ (orange dashed line). Subject to the input configuration $(x_1, x_2) \in \reals^2$, the switch separates left and right half-planes of $\reals^2$, and subsequently selects one of two linear separators (black lines) to correctly distinguish between the upper and lower lattice points (Section~\ref{subsec:NonlinearAdaptiveSwitches}).}
    \label{fig:xor_basic_example}
\end{figure*}
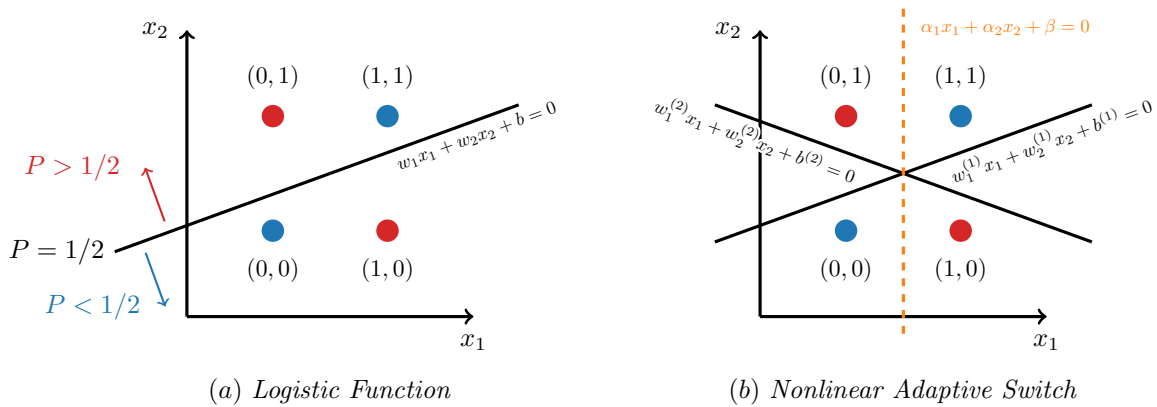



\section{Modeling Discrete Distributions}
\label{sec:xor_example}

The joint distribution of a vector of discrete random variables $X_{[1:n]} \equiv [X_1, X_2,\ldots,X_n]^{T}$ can be decomposed into conditional distributions via the chain rule,
\begin{align}
p_{X_{[1:n]}}(x_{[1:n]}) = \prod_{k=0}^{n-1} p_{X_{k+1}|X_{[1:k]}}\left(x_{k+1} \bigl| x_{[1:k]}\right). \label{eqn:JointDistributionChainRule}
\end{align}
Modeling the joint distribution in~\eqref{eqn:JointDistributionChainRule} implies modeling successive conditional distributions. Each conditional distribution captures higher-order interactions between random variables.


\subsection{Limitations of Sigmoid Functions}\label{subsec:limitations_sigmoid}
A parametric model for each conditional distribution in~\eqref{eqn:JointDistributionChainRule} could be the logistic or sigmoid function. However, this model is too simplistic. Consider the following elementary example. Let $\{X_1, X_2\}$ denote two independent Bernoulli$\,\left(1/2\right)$ random variables. Let $X_3 = X_1 \oplus X_2$ denote a third binary random variable, which introduces higher-order dependencies. The \texttt{XOR} operation $\oplus$ represents addition modulus $2$. Consider a parametric model for the conditional distribution $p_{X_3|X_1,X_2}(x_3 | x_1, x_2)$,
\begin{equation}
  p_{X_3|X_1,X_2}(1 | x_1, x_2) = \sigma(w_1 x_1 + w_2 x_2 + b). \label{eq:sigmoid-xor}
\end{equation}
For this specific example, if parameters $\{w_1, w_2, b\}$ are learned via maximum-likelihood estimation, it is impossible to achieve a likelihood greater than $1/2$.

To see this, note that the logistic form of
(\ref{eq:sigmoid-xor}) represents a linear separator in $\reals^2$ in a
probabilistic sense (Figure~\ref{fig:xor_basic_example}). The half-plane $w_1 x_1 + w_2 x_2 + b > 0$ represents
points $(x_1, x_2)$ that are more likely to generate the event $X_3 = 1$, and the half-plane $w_1 x_1 + w_2 x_2 + b < 0$ represents points $(x_1, x_2)$ that are more likely to generate the event $X_3 = 0$. In order to model the deterministic function $X_3 = X_1 \oplus X_2$ correctly, the linear separator must partition the space of integral lattice points $\{(0,0), (1,1), (1,0), (0,1)\}$ correctly. Illustrated in blue, lattice points $\{(0,0), (1,1)\}$ should imply a higher likelihood of generating the event $X_3 = 0$. Illustrated in red, lattice points $\{(1,0), (0,1)\}$ should imply a higher likelihood of generating the event $X_3 = 1$. However, no linear separator in $\reals^2$ is able to differentiate between the blue and red sets of lattice points.

We note that the insufficiency of a single linear separator has been observed for binary logistic regression for classification tasks, thereby motivating the use of deep neural networks. However, the context of our analysis is the statistical modeling of discrete distributions, by maximizing the likelihood of the data given the model.


\subsection{A Nonlinear, Adaptive Switch}\label{subsec:NonlinearAdaptiveSwitches}
We construct a statistical model which utilizes a nonlinear, adaptive switch. This model is designed to be simple enough to train efficiently. For the example of Figure~\ref{fig:xor_basic_example}, consider \textit{two} linear separators in $\reals^2$. Let $Y_j$, $j \in \{1,2\}$, denote two binary random variables. Their conditional distributions are specified as follows:
\begin{align}
p_{Y_j|X_1, X_2}(1 | x_1, x_2) = \sigma\left(w^{(j)}_1 x_1 + w^{(j)}_2 x_2 + b^{(j)}\right). \notag
\end{align}
Auxiliary variables $Y_j$ have corresponding linear separators in $\reals^2$ which depend on the input configuration $(x_1, x_2)$. By selecting either $Y_1$ or $Y_2$ based on the input configuration $(x_1, x_2)$, it is possible to significantly improve the modeling of $p_{X_3|X_1,X_2}(x_3|x_1,x_2)$. More precisely, consider a discrete (non-binary) random variable $Z$ taking values in $\{1, 2\}$, which represents an adaptive switch. The switch also depends on the input configuration $(x_1, x_2)$. This switch can itself be modeled by a logistic function\footnote{The logistic function is not required for this elementary example of a switch. However, we introduce it to be consistent with the softmax function used to model adaptive switches in multilayer networks.} as follows:
\begin{align}
p_{Z|X_1, X_2}(1 | x_1, x_2) & = \sigma(\alpha_1 x_1 + \alpha_2 x_2 + \beta). \notag
\end{align}
Note that $p_{Z|X_1,X_2}(2 | x_1, x_2) = 1 - p_{Z|X_1, X_2}(1 | x_1, x_2)$. Random variable $Y_Z$ selects either $Y_1$ or $Y_2$ adaptively. An adaptive switch improves upon the limited sigmoid function of~\eqref{eq:sigmoid-xor}, and models the target conditional distribution as follows,
\begin{align}
& p_{X_3|X_1,X_2}(1|x_1,x_2) \notag \\
& \quad \quad = \sum_{j=1,2} p_{Z|X_1, X_2}(j | x_1, x_2) p_{Y_j|X_1,X_2}(1|x_1,x_2). \notag
\end{align}
This switch can achieve a maximum likelihood of nearly $1$ after stochastic gradient optimization. As illustrated in Figure~\ref{fig:xor_basic_example}, there exist three separator lines for this adaptive switch. The slanted black lines indicate the two separators for $Y_j$, $j \in \{1,2\}$, and the dashed orange line indicates the separator for the switch variable $Z$. A likelihood of nearly $1$ is obtained as a result of the switch variable $Z$ correctly distinguishing between the two half-planes corresponding to $X_1 = 0$ and $X_1 = 1$. Within each half-plane, the slanted black lines correctly distinguish between the upper and lower integral lattice points.



\section{Network Architecture}
\label{sec:one-layer-switch-networks}

The example of Section~\ref{subsec:NonlinearAdaptiveSwitches} can be generalized to include switching between multiple linear separators in high-dimensional spaces. Experiments show that the adaptive switch is sufficient for non-trivial representation of discrete data. Furthermore, we devise deep network architectures for learning high-dimensional discrete distributions. Deep switch networks exhibit an improved performance in generative modeling compared to single-layer networks.

\subsection{Single-layer Networks}
To model the distribution of $X_{k+1}$ conditioned on $X_{[1:k]}$, we define $m$ binary auxiliary random variables $\Yk_1, \dots, \Yk_m$. The conditional distribution of these auxiliary random variables is specified in parametric form as follows. For $1 \leq j \leq m$,
\begin{equation}
  \label{eq:onelayer-pYi}
  p\bigl(\Yk_j = 1 \bigl| x_{[1:k]}; \tk \bigl) = \sigma\bigl(x^T_{[1:k]} w_{\Yk_j} + b_{\Yk_j}\bigl).
\end{equation}
Each auxiliary variable represents a linear separator in the probabilistic sense as discussed in Section~\ref{subsec:NonlinearAdaptiveSwitches}. A single-layer switch network chooses between these $m$ linear separators in high-dimensional space, thereby partitioning sets of integral lattice points. In~\eqref{eq:onelayer-pYi}, $\tk$ signifies all model parameters for the $k$-th conditional distribution. The number of parameters to learn for $m$ auxiliary variables for the $k$-th conditional is $\mathcal{O}(m k)$. Specifically, these parameters are 
\begin{align}
w_{\Yk_j} \in \reals^k, ~b_{\Yk_j} \in \reals, \mbox{for}~1 \leq j \leq m. \notag 
\end{align}
Since $0 \leq k \leq n-1$ for all conditionals, the total number of parameters is $\mathcal{O}(m n^2)$. For $m$ a constant, we note that this complexity in parameters to learn is similar to that of a fully-connected single-layer neural network, and similar to that of RBMs with an equivalent number of hidden latent variables. 

\subsection{Statistical Model of Adaptive Switching}
To model the $k$-th conditional, we define a non-binary switch random variable, $\Zk$, which takes values in the set $\{1, \dots, m\}$. The switch random variable selects a particular linear separator out of $m$ separators. The switch value $\Zk$ is determined in an adaptive fashion based on the value of the input variables $X_{[1:k]}$. The switching mechanism is modeled by a softmax distribution. More precisely, for $1 \leq j \leq m$,
\begin{align}
& p\left(\Zk = j \Bigl| x_{[1:k]}; \tk\right) \notag \\
& \quad \quad \quad = \frac{\exp \left(  x_{[1:k]}^T \alpha^{(j)}_{\Zk} + \beta^{(j)}_{\Zk} \right)}{
\displaystyle\sum_{j'=1}^m \exp \left(  x_{[1:k]}^T \alpha^{(j')}_{\Zk}
  + \beta^{(j')}_{\Zk} \right)}. \label{eq:onelayer-pZ}
\end{align}
In~\eqref{eq:onelayer-pZ}, $\tk$ signifies all model parameters for the $k$-th conditional distribution. These parameters include those for the switch defined above, namely, 
\begin{align}
\alpha^{(j)}_{\Zk} \in \reals^k, ~\beta^{(j)}_{\Zk} \in \reals, \mbox{for}~1 \leq j \leq m. \notag 
\end{align}
Thus, the adaptive switch also requires learning of $\mathcal{O}(mk)$ parameters for the $k$-th conditional, and a total of $\mathcal{O}(mn^2)$ parameters for all conditional distributions.

\subsection{Log-Likelihood Objective}
Based on the defined auxiliary and switch random variables, the $k$-th target conditional distribution is modeled by selecting one of the variables $\Yk_1,
\dots, \Yk_m$ based on the value of $\Zk$. More precisely, utilizing the equation terms written in~\eqref{eq:onelayer-pYi} and~\eqref{eq:onelayer-pZ}, 
\begin{align}
& p\bigl(X_{k+1} = 1|x_{[1:k]}; \tk\bigl) = \notag \\
& \sum_{j=1}^m p\bigl(\Zk \!\! = \! j \bigl| x_{[1:k]}; \tk\bigl) p\bigl(\Yk_j \!\! = \!\! 1 \bigl| x_{[1:k]}; \tk\bigl). \label{eq:onelayer-Xtilde}
\end{align}
We optimize the model parameters $\theta^{(1)}, \ldots, \theta^{(n-1)}$ in order to maximize the empirical log likelihood. More precisely, consider $I$ data samples $\vx^{(1)}, \dots, \vx^{(I)}$ obtained from the true distribution. Each individual data sample is $n$-dimensional, $\vx^{(i)} = (x^{(i)}_1, \dots, x^{(i)}_n)$. For $0 \leq k \leq n-1$, we optimize for the following empirical log-likelihood, 
\begin{equation}
  \label{eq:onelayer-Lk}
  L^{(k)} := \frac{1}{I} \sum_{i=1}^I \log p\left(X_{k+1} = x^{(i)}_{k+1} \bigl| x^{(i)}_{[1:k]}; \tk\right).
\end{equation}
The conditional probabilities have the specific form given in~\eqref{eq:onelayer-Xtilde}, and the log-likelihood is differentiable. Thus, stochastic gradient descent allows for efficient training. Note that the optimization can be done in a distributed fashion in parallel. The objectives $L^{(1)}, \dots, L^{(n-1)}$ can be optimized independently in order to find the optimal parameters $\theta^{(1)}, \dots, \theta^{(n-1)}$. 

\section{Multilayer Switch Networks}
\label{sec:two-layer-switch}

\begin{figure}
  \centering
  \includegraphics[width=\linewidth]{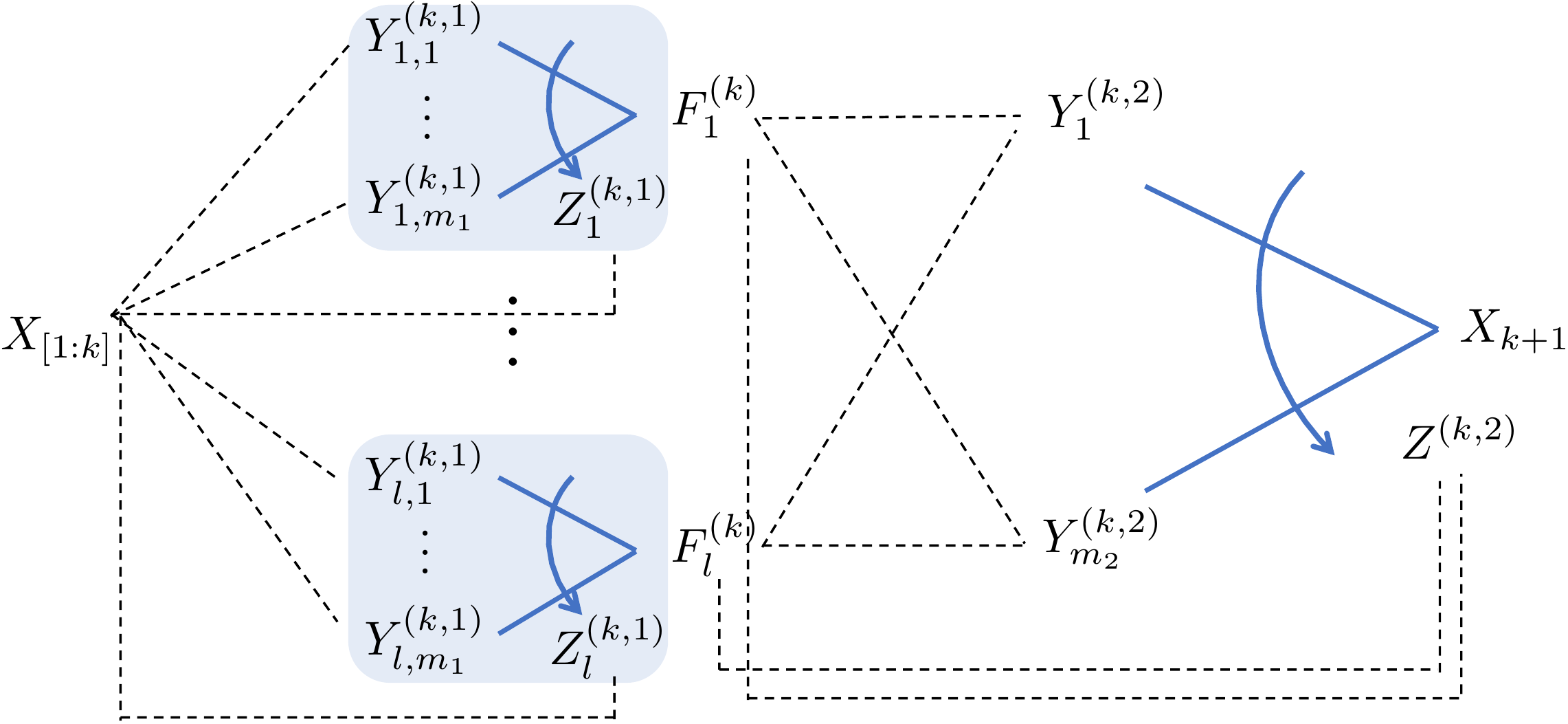}
  \caption{\textit{The architecture of two-layer switch networks:} We have a total of       $l$ intermediate variables $F^{(k)}_1, \dots, F^{(k)}_l$, each of which is
    the output of a single-layer switch network. Subsequently, these variables are fed     into another single-layer switch network to generate the final output $X_{k+1}$.
    Dotted lines indicate dependencies between variables.}
  \label{fig:switch-2-layer-architecture}
\end{figure}

In this section, we propose a deep architecture, extending the single-layer adaptive switch described in Section~\ref{sec:one-layer-switch-networks}. To avoid excessive mathematical notation, we focus on the architecture of a two-layer switch network, depicted in Figure~\ref{fig:switch-2-layer-architecture}. Generalization to more layers follows inductively and naturally by replicating switch network primitives.

\subsection{Combining Switch Networks}
To better model the $k$-th conditional distribution of discrete random variables, we first replicate and instantiate $l$ single-layer switch networks, as described in Section~\ref{sec:one-layer-switch-networks}. Each switch network utilizes $m_1$ auxiliary random variables. More precisely, the $i$-th switch network contains $m_1$ auxiliary variables denoted by $\bigl\{\Yki_{i,1}, \dots, \Yki_{i,m_1}\bigl\}$, and a non-binary switch random variable denoted by $\Zki_i$. For indices $1 \leq i \leq l$ and $1 \leq j \leq m_1$, the conditional distributions of the auxiliaries are
 specified as logistic functions,
\begin{equation}
  \label{eq:two-layer_Y1}
  p\bigl(\Yki_{i,j} \! \! = \! 1 \bigl| x_{[1:k]}; \tk_1 \bigl) = \sigma \bigl(x_{[1:k]}^T w_{\Yki_{i,j}} \! + b_{\Yki_{i,j}} \bigl).
\end{equation}
The conditional distribution for the adaptive switch for the $i$-th single-layer network is specified by a softmax function,
\begin{align}
\label{eq:twolayer-pZ}
& p\Bigl(\Zki_i = j \bigl | x_{[1:k]}; \tk_1 \Bigl) \notag \\
& \quad \quad \quad \quad = \frac{\exp \Bigl(  x_{[1:k]}^T \alpha^{(j)}_{\Zki_i} + \beta^{(j)}_{\Zki_i} \Bigl)}{\displaystyle\sum_{j'=1}^{m_1} \exp \left(  x_{[1:k]}^T \alpha^{(j')}_{\Zki_i} + \beta^{(j')}_{\Zki_i} \right)}.
\end{align}
The model parameters for the first layer are denoted by $\tk_1$, and consist of the following coefficients,
\begin{align}
& w_{\Yki_{i,j}} \in \reals^{k}, ~b_{\Yki_{i,j}} \in \reals, \notag \\
& \alpha^{(j)}_{\Zki_i} \in \reals^k, \beta^{(j)}_{\Zki_i} \in \reals, \notag
\end{align}
for $1 \leq i \leq l$, $1 \leq j \leq m_1$. The total number of parameters is $\mathcal{O}(l m_1 k)$ for the $k$-th conditional, and $\mathcal{O}(l m_1 n^2)$ for all conditionals.

\subsection{Intermediate Variables Between Layers}
As shown in Figure~\ref{fig:switch-2-layer-architecture}, the outputs of the $l$ single-layer networks are labeled by variables $\Fk_1, \dots, \Fk_l$. These outputs are called intermediary variables prior to the second layer. The conditional distributions of these intermediary variables are specified as follows,
\begin{align}
& p\bigl(\Fk_i = 1 | x_{[1:k]}; \tk_1\bigl) = \sum_{j=1}^{m_1} p\bigl(\Zk_i = j | x_{[1:k]}; \tk_1\bigl) \times \notag \\
& ~~~~~~~~~~~~~~~~~~~~~~~~~~~~~ p\bigl(\Yki_{i,j} = 1|x_{[1:k]}; \tk_1\bigl). \label{eq:twolayer-pFi}
\end{align}

We assume that the $\Fk_i$, $1 \leq i \leq l$ are conditionally independent given $X_{[1:k]}$. More precisely, for any configuration of the intermediate variables $\fk_{[1:l]}
\in \{0,1\}^l$, we assume the following product decomposition,
\begin{equation*}
  p\bigl(\fk_{[1:l]}| x_{[1:k]}; \tk_1\bigl) = \prod_{i=1}^l p\bigl(\fk_i| x_{[1:k]}; \tk_1\bigl).
\end{equation*}

\subsection{Processing of the Second Layer}
Having constructed the intermediate variables $\Fk_1, \dots, \Fk_l$, the second and final output layer is once again an adaptive switch network. This switch network is modeled in parametric form with $m_2$ auxiliary random variables $\bigl\{ \Ykii_1, \dots, \Ykii_{m_2} \bigl\}$ and switch random variable $\Zkii$. More precisely, for $1 \leq j \leq m_2$, conditioned on any configuration of the intermediate variables $\fk_{[1:l]} \in \{0,1\}^l$,
\begin{equation}
\label{eq:two-layer-Yk2}
p\bigl(\Ykii_j \! = 1 \bigl| \fk_{[1:l]}; \tk_2 \bigl) = \sigma \bigl((\fk_{[1:l]})^T w_{\Ykii_j} \! + b_{\Ykii_j} \bigl).
\end{equation}
The conditional distribution for the adaptive switch of the second layer is given by a softmax function,
\begin{align}
\label{eq:twolayer-Z2}
& p\Bigl(\Zki_i = j \bigl | x_{[1:k]}; \tk_1 \Bigl) \notag \\
& \quad \quad \quad \quad = \frac{\exp \left(  (\fk_{[1:l]})^T \alpha^{(j)}_{\Zkii} + \beta^{(j)}_{\Zkii} \right)}{\displaystyle\sum_{j'=1}^{m_2} \exp \left(  (\fk_{[1:l]})^T \alpha^{(j')}_{\Zkii} + \beta^{(j')}_{\Zkii} \right)}.
\end{align}
The model parameters for the second layer are denoted by $\tk_2$, and consist of the following coefficients,
\begin{align}
& w_{\Ykii_j} \in \reals^l, ~b_{\Ykii_j} \in \reals, \notag \\
& \alpha^{(j)}_{\Zkii} \in \reals^l, ~\beta^{(j)}_{\Zkii} \in \reals, \notag
\end{align}
for $1 \leq j \leq m_2$. The total number of parameters is $\mathcal{O}(m_2 l)$ for the $k$-th conditional, and $\mathcal{O}(m_2 l n)$ for all conditionals. Thus, a two-layer switch network can be characterized by the triplet $(m_1, l, m_2)$ which specifies the number of parameters in both layers.

\subsection{Log-Likelihood Objective}
The target distribution of $X_{k+1}$ conditioned on the intermediate variables is governed by
\begin{align}
& p\bigl(X_{k+1} = 1 |\fk_{[1:l]}; \tk_2\bigl) = \sum_{j=1}^{m_2} p\bigl(\Ykii_j = 1| \fk_{[1:l]}; \tk_2\bigl) \times \notag \\
& ~~~~~~~~~~~~~~~~~~~~~~~~~~~~~~ p\bigl(\Zkii = j | \fk_{[1:l]}; \tk_2\bigl). \label{eq:twolayer_Xk+1}
\end{align}
The empirical log-likelihood of the $k$-th conditional given $I$ discrete samples $\vx^{(1)}, \dots, \vx^{(I)}$ is given by,
\begin{equation}
  \label{eq:twolayer_Lk}
  \begin{aligned}
    L^{(k)} &:= \frac{1}{I} \sum_{i=1}^I \log p\bigl(X_{k+1} = x^{(i)}_{k+1} | x^{(i)}_{[1:k]}; \tk \bigl) \\
    &=  \frac{1}{I} \sum_{i=1}^I \log \sum_{\fk_{[1:l]}} p\bigl(\fk_{[1:l]}| x^{(i)}_{[1:k]}; \tk_1 \bigl) \times \\
    &\qquad \qquad p\bigl(X_{k+1} = x^{(i)}_{k+1} | \fk_{[1:l]}; \tk_2 \bigl).
  \end{aligned}
\end{equation}

\begin{table*}[t]
  \centering
{\footnotesize
\begin{tabular}{@{}lllll@{}}
\toprule
Model                                & Epochs trained & -(Log-likelihood) & TV distance & JS distance \\ \midrule
One layer, $m=4$                     & 2910           & 6.887280       & 0.303341                 & 0.020051                 \\
One layer, $m=16$                    & 3120           & 6.833584       & 0.156232                 & 0.006034                 \\
  Two layer, $(m_1, l, m_2) = (4,4,8)$ & 1090           & 6.82794115     & 0.13848737               & 0.00460584               \\
  Two layer, $(m_1, l, m_2) = (2, 8, 32)$ & 2320 & 6.820030 & 0.102167 & 0.002584 \\
  \bottomrule
\end{tabular}
}
\caption{\label{tab:synthetic_tv-js} Comparing the performance of switch network models for a synthetic discrete distribution. The log-likelihood is computed using an expectation over the true distribution. Total variation distance (TV) and Jensen--Shannon distance (JS) between the learned distributions and the true distribution are computed exactly.}
\end{table*}

\vspace{-5mm}
\subsection{MCMC Gradient Approximations}
The log-likelihood given in~\eqref{eq:twolayer_Lk} contains a summation $\sum_{\fk_{[1:l]}}$ over all possible configurations of the intermediate variables $\Fk_{[1:l]}$ in $\{0,1\}^l$. In two-layer experiments, we set $l = 4$ or $l = 8$ which produces reasonable results. The log-likelihood is differentiable and stochastic gradient descent (SGD) can be applied to the objective. However, in general, in order to overcome the computational difficulty in computing this objective and its gradients due to the exponential number of configurations of the intermediate variables when $l$ is large, we discuss methods in the supplement to approximate the gradient of $L^{(k)}$. Specifically, Monte-Carlo Markov Chain (MCMC) gradient approximations are derived which provide a \textit{trade-off} in the desired accuracy of approximation versus computational complexity.

\begin{figure}
\begin{center}
  \input{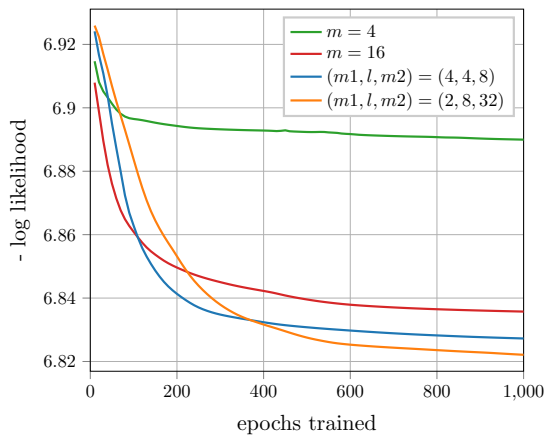}
  \caption{  \label{fig:synthetic-epoch1000}
    Log-likelihood (LL) curves of single-layer and multi-layer switch networks after SGD training over many epochs, on data samples from a synthetic distribution. The optimal LL value is $6.81$.}
  \end{center}
\end{figure}

\section{Experimental Results}
\label{sec:experimental}

We evaluate single-layer and multilayer switch network architectures, as described in Sections \ref{sec:one-layer-switch-networks} and \ref{sec:two-layer-switch}, by conducting experiments on diverse sets of data.



\subsection{Synthetic Discrete Distribution}\label{subsec:synthetic_discrete_distrib}

We generated a synthetic distribution on $n = 10$ binary variables $X_{[1:10]}$,
where the probability of each of the $2^n$ possible configurations is proportional to a uniform random number in the interval $[0.1,1]$. The probabilities were normalized to ensure a valid probability distribution. Then,  $10^5$ i.i.d. samples were generated from this distribution as the training data.


\subsubsubsection{\bf{Training Setup:}} The generated database was used to train several instances of the single-layer and two-layer switch networks. More precisely, two instances of a one-layer network with parameters $m = 4$ and $m=16$, and two instances of a two-layer network with parameters $(m_1, l, m_2) = (4,4,8)$ and $(m_1, l, m_2) = (2, 8, 32)$ were trained. We used stochastic gradient descent (SGD) to optimize for the log-likelihood objectives in~\eqref{eq:onelayer-Lk} and~\eqref{eq:twolayer_Lk}, using batch size $1000$ and learning rate $10$. For the two-layer objective, we optimized directly for the exact likelihood in~\eqref{eq:twolayer_Lk} without employing an approximation, since $l$ was not too large to cause computational difficulties.


\subsubsubsection{\bf{Training Log-likelihood Performance:}} Figure~\ref{fig:synthetic-epoch1000} illustrates the empirical (negative) log-likelhood,
i.e. $-\sum_{k=0}^{n-1} L^{(k)}$, after $1000$ training epochs. Note that due to the
minus sign, we should minimize the objective. It is easy to see that, using Gibbs' inequality, the value of the objective cannot go below the entropy of the empirical distribution of the database, which is approximately $6.81$ in this example. As shown in Figure~\ref{fig:synthetic-epoch1000}, the likelihood performance improves by increasing the hyper-parameter $m$ for single-layer architectures, and also by updating from one-layer to two-layer structures. The two-layer network with $(m_1, l, m_2) = (2,8,32)$ nearly achieves the
optimal value of $6.81$.


\subsubsubsection{\bf{Generalization Ability:}} Note that since the ground truth synthetic distribution is known, we can compare the joint distribution learned by our model using \eqref{eq:onelayer-Xtilde} and \eqref{eq:twolayer_Xk+1}, and compare it with the ground truth. For the network architectures described in our training setup, we compared the total variation (TV) distance and Jensen--Shannon (JS) distance between the learned distributions and the ground truth. Table~\ref{tab:synthetic_tv-js} illustrates these metrics. The TV and JS distances decrease as $m$ increases, and also if the model is updated to a two-layer architecture.


\begin{figure}[t]
  \begin{center}
    \input{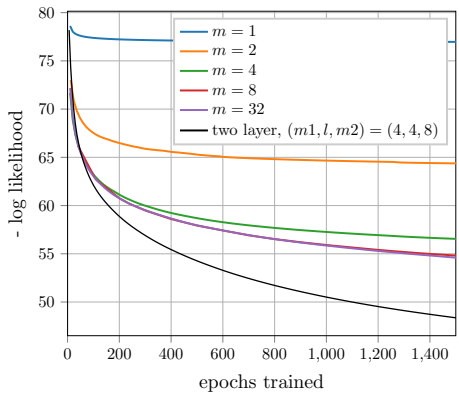}
    \caption{\label{fig:mnist_one-two-layer} Empirical log-likelihood performance of training on a binary version of the \texttt{MNIST} handwritten digits dataset for different configurations of single-layer and two-layer switch networks. Single-layer architectures do not exhibit improvement beyond $m=8$. However, the two-layer network structure provides an immense improvement over single-layer networks. These log-likelihood results may be compared to baseline results in the literature~\cite{Tieleman2008}.}
  \end{center}
\end{figure}

\subsection{High-Dimensional Binary \texttt{MNIST}}\label{sec:high_dimensional_binary_MNIST}

We converted each data sample of the \texttt{MNIST} dataset of handwritten digits into a discrete binary sequence. In its original form, each sample is a $28 \times 28$ image consisting of pixels which have values in the interval $[0,255]$. We converted each pixel
to a bit by treating values above $150$ as $1$ and values below $150$ as $0$. We then concatenated the rows of each image from top to bottom to form a binary vector of length $n = 784$. Thus, each data sample is represented by a high-dimensional binary vector in $\{0, 1\}^{784}$. This binary version of the \texttt{MNIST} dataset does not have an entropy of $784$ bits. The handwritten variations occurring in the dataset likely constitute less than $50$ bits of entropy.


\subsubsubsection{\bf{Training Setup:}} We trained the discretized \texttt{MNIST} dataset using a one-layer switch network with the value of the hyperparameter $m$ ranging in the set  $\{1,2,4,8,32\}$. A single-layer network with $m=1$ is essentially a logistic conditional model similar to that of~\eqref{eq:sigmoid-xor}. Also, we trained a two layer network with $(m_1, l, m_2) = (4,4,8)$. We used SGD for optimizing the log-likelihood in
\eqref{eq:onelayer-Lk} for the one layer architecture, and also the exact likelihood
objective in \eqref{eq:twolayer_Lk}. In both cases, we used a batch size of $300$ and
learning rate $10$.

\subsubsubsection{\bf{Training Log-Likelihood Performance:}} Figure~\ref{fig:mnist_one-two-layer} illustrates minus the training log likelihood objective, i.e. $-\sum_{k=0}^{n-1} L^{(k)}$, for the above configurations of the one layer and the two layer network. Increasing $m$ from 1 to 2, i.e. adding one more logistic variable, greatly improves the likelihood performance. Increasing $m$ above $8$ does not improve the training likelihood. However, the two layer configuration exhibits a significant improvement in the training likelihood.


\begin{figure}[t]
  \centering
  \includegraphics[width=0.82\linewidth]{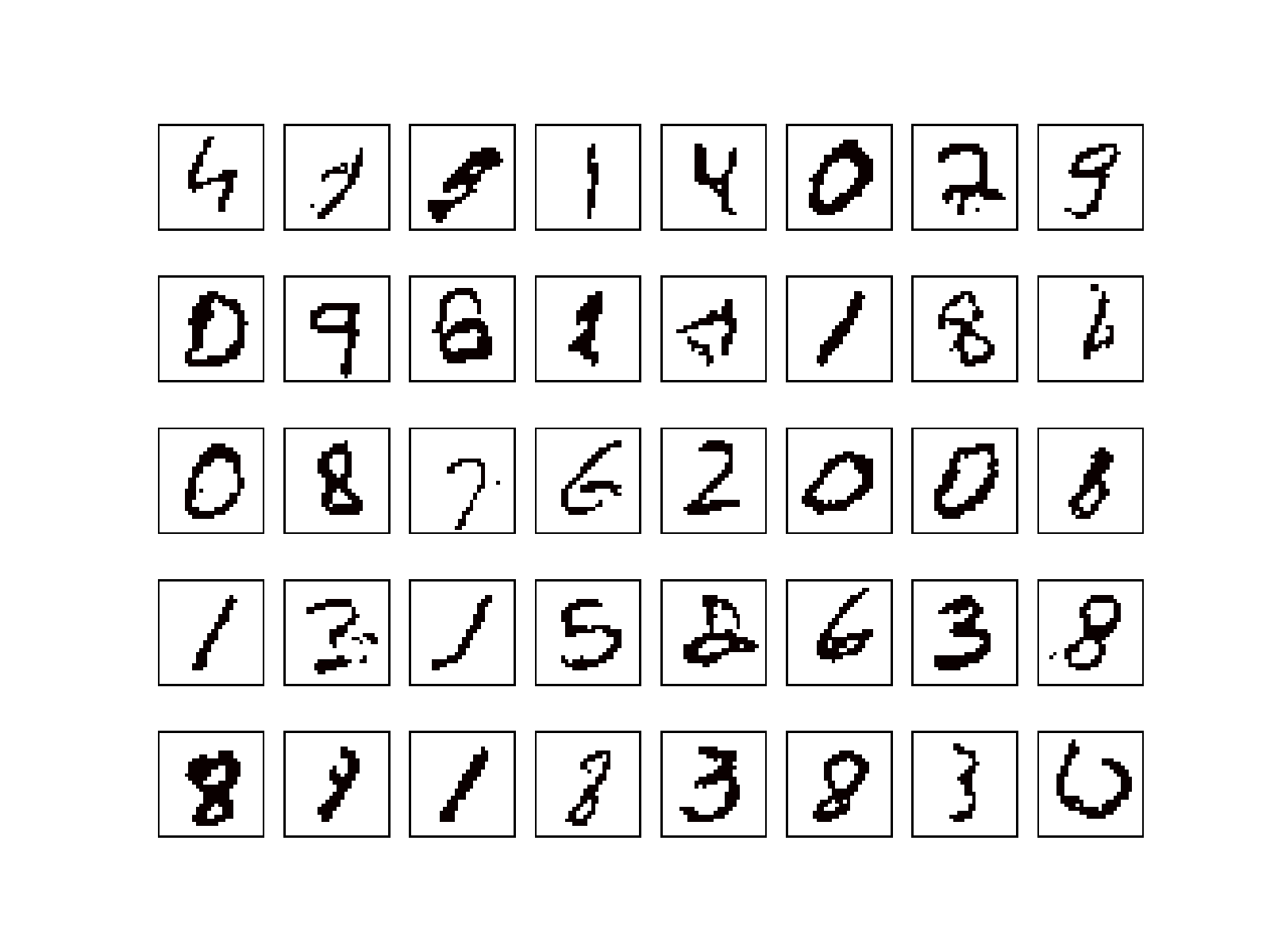}
  \caption{Generated binary digits from a simple 2--layer switch network with parameters $(m_1, l, m_2) = (4,4,8)$ after approximately 1500 epochs of training. }
  \label{fig:two-layer-mnist-samples}
\end{figure}

\begin{figure}[t]
\begin{center}
\begin{tikzpicture}[scale=0.62, trim axis left, trim axis right]

\definecolor{color0}{rgb}{0.12156862745098,0.466666666666667,0.705882352941177}

\begin{axis}[
  ybar,
  width = 1.2\linewidth,
tick align=outside,
tick pos=left,
grid = both,
x grid style={lightgray!92.02614379084967!black},
xlabel={$m$},
symbolic x coords={1, 2, 4, 8},
xtick = {1,2,4,8},
y grid style={lightgray!92.02614379084967!black},
ylabel={- log likelihood (test)},
x label style = {scale=1.2},
y label style = {scale=1.2},
ymin=72.1890699106981, ymax=79.0992601872595
]
\addplot 
table [row sep=\\]{%
1	78.785160629234 \\
2	72.5031694687236 \\
4	73.2756491641709 \\
8	75.7667167737573 \\
};
\end{axis}

\end{tikzpicture}
  \caption{Test likelihood performance versus model complexity for single-layer switch networks trained on binary \texttt{MNIST} data.}
  \label{fig:mnist_bias-variance}
  \end{center}
\end{figure}
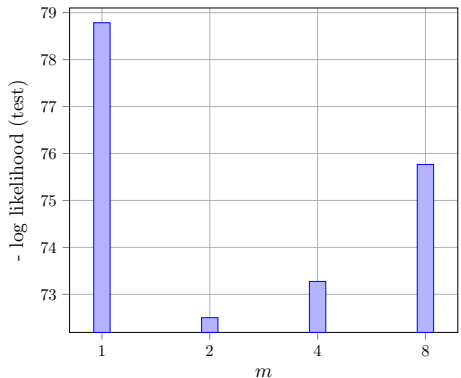

\subsubsubsection{\bf{Generalization Ability:}} Figure~\ref{fig:two-layer-mnist-samples} illustrates samples generated by a 2--layer switch network with parameters $(m_1, l, m_2) = (4,4,8)$. We also computed the test error performance for single-layer networks on the \texttt{MNIST} test dataset which consists of $10,000$ test samples. Figure~\ref{fig:mnist_bias-variance} illustrates the test log-likelihood performance versus the model complexity (i.e. the hyper-parameter $m$) for single-layer switch networks. This plot obeys the general bias-variance trade-off.



\subsection{Web-Crawled Language Data}

We crawled the \href{mit.edu}{web.mit.edu} website and extracted words of length up to $8$ characters. This dataset consists of $65,568$
words in total, and $686$ distinct words. We processed each character as lowercase, implying $27$ possibilities for characters, including a space and characters $\{ \verb+a+, \dots, \verb+z+ \}$. Each character was converted into an
integer in the set $\{0,\dots 26\}$ by mapping a space to $0$, \verb+a+ to $1$,
\verb+b+ to 2, and so forth. We concatenated the $5$-bit binary representation of the integers corresponding to each character to form a binary sequence of length $n = 8
\times 5 = 40$. Thus, each word was mapped to a high-dimensional binary vector in $\{0, 1\}^{40}$.


\subsubsubsection{\bf{Training Setup:}} We trained two instances of the single-layer switch network with parameters $m=4$
and $m=16$ and one instance of the two layer switch network with $(m_1, l, m_2)
= (4,4,8)$. For the single-layer configurations, we optimize the objective as in \eqref{eq:onelayer-Lk}, and for the two layer configuration, we optimize the exact objective as in
\eqref{eq:twolayer_Lk}, using SGD.

\begin{figure}[t]
  \begin{center}
\begin{tikzpicture}[scale=0.72]


\definecolor{color1}{rgb}{1,0.498039215686275,0.0549019607843137}
\definecolor{color0}{rgb}{0.12156862745098,0.466666666666667,0.705882352941177}
\definecolor{color2}{rgb}{0.172549019607843,0.627450980392157,0.172549019607843}

\begin{axis}[
  width = 1.2\linewidth,
legend cell align={left},
legend entries={{1--layer, $m = 4$, LR = 10, BS = 30},{1--layer, $m = 16$, LR = 10, BS = 50},{2--layer, $(m1,l,m2) = (4,4,8)$, LR = 10, BS = 50}},
legend style={draw=white!80.0!black, very thick, nodes={scale=0.8, transform shape}},
tick align=outside,
tick pos=left,
grid=both,
x grid style={lightgray!92.02614379084967!black},
x label style = {scale=1.2},
y label style = {scale=1.2},
xlabel={epochs trained},
xmin=10, xmax=200,
y grid style={lightgray!92.02614379084967!black},
ylabel={- log likelihood},
ymin=5.469285267622, ymax=7.49891787194108
]
\addlegendimage{no markers, color0}
\addlegendimage{no markers, color1}
\addlegendimage{no markers, color2}
\addplot [very thick, color0]
table [row sep=\\]{%
10	7.40666184447203 \\
20	7.25030674056917 \\
30	7.20142854646279 \\
40	7.15254140505401 \\
50	7.12078591897538 \\
60	7.09541631246976 \\
70	7.07972523162614 \\
80	7.06881044436468 \\
90	7.05952181055177 \\
100	7.05327551421297 \\
110	7.04796077638994 \\
120	7.0435828292707 \\
130	7.03978554115609 \\
140	7.03599002470516 \\
150	7.03214376020225 \\
160	7.02782061782159 \\
170	7.0219693401811 \\
180	7.01842998972852 \\
190	7.01387737908079 \\
200	7.01112916484 \\
210	7.00891605311049 \\
};
\addplot [very thick, color1]
table [row sep=\\]{%
10	6.84702099389731 \\
20	6.60599902614515 \\
30	6.47640525025751 \\
40	6.4008985401362 \\
50	6.35408244706229 \\
60	6.32433247780084 \\
70	6.30190386461861 \\
80	6.28481843246661 \\
90	6.27209400101416 \\
100	6.26160209647629 \\
110	6.24417534611834 \\
120	6.2308596273157 \\
130	6.22165723212959 \\
140	6.21349096004704 \\
150	6.20539366784213 \\
160	6.19767461959366 \\
170	6.19206875593281 \\
180	6.18750860518739 \\
190	6.18327833175049 \\
200	6.17946793311916 \\
210	6.17612775704306 \\
220	6.17307554668632 \\
230	6.17019274204049 \\
240	6.16397823985764 \\
250	6.15998703143789 \\
260	6.15727231289626 \\
270	6.15481422690517 \\
280	6.15235488401777 \\
290	6.14683891591594 \\
300	6.14383618400347 \\
310	6.14144953535686 \\
320	6.13955458665968 \\
330	6.13787274201001 \\
340	6.13627920156823 \\
350	6.13485667894775 \\
360	6.13339252810707 \\
370	6.13206701452406 \\
380	6.13084946802794 \\
390	6.12968126014421 \\
400	6.1285608703626 \\
410	6.1274577559477 \\
420	6.12631619871036 \\
430	6.12409187849573 \\
440	6.12260064964773 \\
450	6.12044290752056 \\
460	6.11866781376 \\
470	6.11731188891817 \\
480	6.11581722288622 \\
};
\addplot [very thick, color2]
table [row sep=\\]{%
10	6.19633408432128 \\
20	5.92221908166539 \\
30	5.8294517424074 \\
40	5.779460323698 \\
50	5.74805970001034 \\
60	5.72689658276795 \\
70	5.71215753481374 \\
80	5.69586945437186 \\
90	5.68098167679273 \\
100	5.66829053519905 \\
110	5.65323749781965 \\
120	5.64609201083658 \\
130	5.64169525124453 \\
140	5.63751450672135 \\
150	5.63226820312047 \\
160	5.62944943456387 \\
170	5.62550108047071 \\
180	5.62072895155143 \\
190	5.6151182080248 \\
200	5.60864233596658 \\
210	5.60461637137269 \\
220	5.60007628974563 \\
230	5.60067701931621 \\
240	5.59871634187584 \\
250	5.59740542840518 \\
260	5.59543140838105 \\
270	5.59228198199889 \\
280	5.58978193519943 \\
290	5.58923040711852 \\
300	5.58583899132827 \\
310	5.58172153815394 \\
320	5.58357115371837 \\
330	5.58349037783046 \\
340	5.58370166224086 \\
350	5.58611296739036 \\
360	5.59092205049274 \\
370	5.58883655467798 \\
380	5.59081715760294 \\
390	5.59090898278555 \\
400	5.59119834465855 \\
410	5.58990446602729 \\
420	5.58781869086124 \\
430	5.58752204045049 \\
440	5.58717583768521 \\
450	5.58509895944007 \\
460	5.58414051787258 \\
470	5.58313974488738 \\
480	5.58228852034426 \\
490	5.58167054159821 \\
500	5.58121609035697 \\
510	5.58097945946338 \\
520	5.58070844544454 \\
530	5.58106331461477 \\
540	5.58108138072748 \\
550	5.58009198161017 \\
560	5.58050123930025 \\
570	5.58112805305245 \\
580	5.57911767073347 \\
590	5.57682035192556 \\
600	5.57349041954512 \\
610	5.57286328618011 \\
620	5.57325000945639 \\
630	5.57293191199369 \\
640	5.57268461260082 \\
650	5.57234095538843 \\
660	5.5723864241354 \\
670	5.57236960099817 \\
680	5.57285494348434 \\
690	5.5719838408877 \\
700	5.56918495893569 \\
710	5.56776569058366 \\
720	5.56619581116502 \\
730	5.56524449496283 \\
740	5.56446692525878 \\
750	5.56328479076547 \\
760	5.56335549465621 \\
770	5.56331950417916 \\
780	5.56362752573295 \\
790	5.5631708906003 \\
800	5.56249171981744 \\
810	5.56303307366852 \\
820	5.56324773502274 \\
830	5.56277542227599 \\
840	5.56174183719759 \\
850	5.56154129509105 \\
860	5.56341175034686 \\
870	5.56330867984707 \\
880	5.56394164325866 \\
890	5.56459085606093 \\
900	5.56491429081507 \\
910	5.56542712184728 \\
920	5.56564066081501 \\
930	5.56619517403396 \\
};
\end{axis}

\end{tikzpicture}
  \caption{Log-likelihood performance of training for learning character-level (i.e., characters encoded into bits) generation of web-crawled words. LR $=$ learning rate, and BS $=$ batch size. Hyper-parameter search was conducted to achieve improved log-likelihoods. Log-likelihoods provide a direct assessment of the training performance of the language model.
}
  \label{fig:mit-likelihood}
\end{center}
\end{figure}
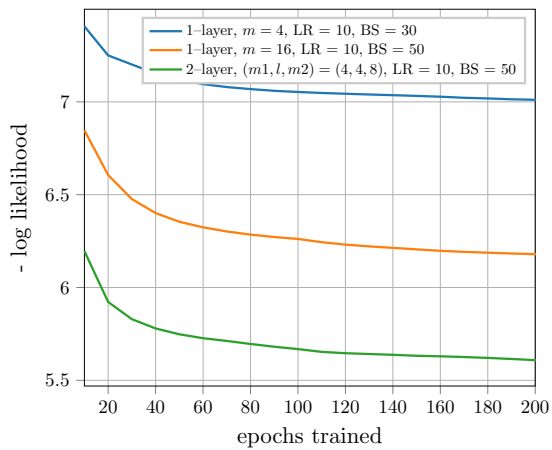

\subsubsubsection{\bf{Training Log-Likelihood Performance}} Figure~\ref{fig:mit-likelihood}  illustrates the likelihood performance for the above mentioned configurations. Note that the two layer architecture which has roughly $m_1 \times l = 16$ many parameters achieves a better likelihood compared to the one layer architecture with roughly the same $m=16$ number of parameters.

\subsubsubsection{\bf{Generalization Ability:}} Figure~\ref{fig:mit-sample-words} illustrates representative words generated using three different network configurations. The two-layer network model generates several recognizable words, with the exception of $\texttt{srereou}$, which nevertheless still follows the vowel and consonant structure of language.


\definecolor{bad}{rgb}{0.78, 0.398, 0}

\definecolor{good}{rgb}{0, 0.1992, 0.398}

\begin{figure}[t]
  \centering
  \tiny

  \begin{subfigure}[t]{0.962\linewidth}
    \begin{center}
    {       \fontfamily{pcr}\selectfont
      \small
    \begin{tabular}{|c|c|c|c|}\hline
      { ~ } & { ~ } & { ~ } \\
      {\textbf{$1$-layer}} & {\textbf{$1$-layer}} & {\textbf{$2$-layer}} \\
      {\textbf{$m = 4$}}  & {\textbf{$m = 16$}} & {\textbf{$(m_1, l, m_2)$}} \\
      {\textbf{$~$}}  & {\textbf{$~$}} & {\textbf{$ ~= (4, 4, 8)$}} \\ \hline
      { $\mbox{LR} = 10$ }  & { $\mbox{LR} = 10$ } & { $\mbox{LR} = 10$ } \\ \hline
      { $\mbox{BS} = 30$ } & { $\mbox{BS} = 50$ } & { $\mbox{BS} = 50$ } \\ \hline
      { $200~\mbox{epochs}$ } & { $240~\mbox{epochs}$ }  &  { $220~\mbox{epochs}$ }  \\ \hline
      { ~ } & { ~ } & { ~ } \\
      {\color{bad}\textbf{aieed}}   & {\color{good}\textbf{a}}     & {\color{good}\textbf{a}} \\
      {\color{good}\textbf{in}}    & {\color{good}\textbf{the}}  & {\color{good}\textbf{alumni}} \\
      {\color{good}\textbf{research}}   & {\color{bad}\textbf{elmc}} & {\color{good}\textbf{we}} \\
      {\color{bad}\textbf{riuu}}    & {\color{good}\textbf{the}}  &{\color{good} \textbf{the}} \\
      {\color{good}\textbf{and}}   & {\color{good}\textbf{of}}     & {\color{good}\textbf{function}} \\
      {\color{good}\textbf{the}}    & {\color{good}\textbf{the}}  &{\color{good} \textbf{help}} \\
      {\color{bad}\textbf{mealeted}}   & {\color{bad}\textbf{robou}}     & {\color{good}\textbf{research}} \\
      {\color{bad}\textbf{roc}}    & {\color{bad}\textbf{syr?kt}}  &{\color{good} \textbf{its}} \\
      {\color{good}\textbf{the}}   & {\color{bad}\textbf{mahh}}     & {\color{good}\textbf{true}} \\
      {\color{good} \textbf{of}}    & {\color{bad}\textbf{nev}}  & {\color{good} \textbf{know}} \\
      {\color{bad}\textbf{vishimn}}  & {\color{good}\textbf{and}} & {\color{bad}\textbf{srereou}}\\
      {\color{bad}\textbf{saydese}}   & {\color{good}\textbf{to}}    & {\color{good}\textbf{world}} \\
      { ~ } & { ~ } & { ~ } \\ \hline
      { ~ } & { ~ } & { ~ } \\
      {\textbf{48/100}}  & {\textbf{74/100}} & {\textbf{91/100}} \\ \hline
    \end{tabular}
  }
  \end{center}

  \end{subfigure}%

  \vspace{5mm}
  \caption{Sample words generated by single-layer and two-layer switch networks.
Correctly generated words are colored in dark blue, and non-dictionary words are colored in orange. LR $=$ learning rate, and BS $=$ batch size. The final row indicates the ratio of the number of dictionary words to the total number of words generated per model. }
  \label{fig:mit-sample-words}
\end{figure}



\section{Conclusion}\label{sec:Conclusion}

We have proposed deep switch networks for learning high-dimensional discrete distributions using maximum-likelihood optimization. The adaptive switch model is simple and interpretable. Comparisons of performance for several network structures were provided. Experimental results were obtained for diverse, discrete distributions.



\subsubsection*{Acknowledgements}
The authors' research was conducted in part at the Technicolor AI Lab in Palo Alto, CA during the summer of 2018. The first author's research is supported by the NSF Science and Technology Center grant CCF--0939370 ``Science of Information''.

\clearpage
\newpage


\bibliography{naveenBIB}

\begin{thebibliography}{}

\bibitem[Arjovsky et~al., 2017]{arjovsky2017wasserstein}
Arjovsky, M., Chintala, S., and Bottou, L. (2017).
\newblock Wasserstein generative adversarial networks.
\newblock In {\em International Conference on Machine Learning}, pages
  214--223.

\bibitem[Arora et~al., 2017]{pmlr-v70-arora17a}
Arora, S., Ge, R., Liang, Y., Ma, T., and Zhang, Y. (2017).
\newblock Generalization and equilibrium in generative adversarial nets
  ({GAN}s).
\newblock In Precup, D. and Teh, Y.~W., editors, {\em Proceedings of the 34th
  International Conference on Machine Learning}, volume~70 of {\em Proceedings
  of Machine Learning Research}, pages 224--232, International Convention
  Centre, Sydney, Australia. PMLR.

\bibitem[Arora et~al., 2018]{arora2018do}
Arora, S., Risteski, A., and Zhang, Y. (2018).
\newblock Do {GAN}s learn the distribution? some theory and empirics.
\newblock In {\em International Conference on Learning Representations}.

\bibitem[Bengio et~al., 2003]{JLMRBengio_2003}
Bengio, Y., Ducharme, R., Vincent, P., and Janvin, C. (2003).
\newblock A neural probabilistic language model.
\newblock {\em Journal of Machine Learning Research}, 3:1137--1155.

\bibitem[Blei et~al., 2017]{Blei_Variational_2017}
Blei, D.~M., Kucukelbir, A., and McAuliffe, J.~D. (2017).
\newblock Variational inference: A review for statisticians.
\newblock {\em Journal of the American Statistical Association},
  112(518):859--877.

\bibitem[Che et~al., 2017]{MaliGAN_2017}
Che, T., Li, Y., Zhang, R., Hjelm, R.~D., Li, W., Song, Y., and Bengio, Y.
  (2017).
\newblock Maximum-likelihood augmented discrete generative adversarial
  networks.
\newblock {\em CoRR}, abs/1702.07983.

\bibitem[Chung et~al., 2015]{NIPS2015_5653_Chung}
Chung, J., Kastner, K., Dinh, L., Goel, K., Courville, A.~C., and Bengio, Y.
  (2015).
\newblock A recurrent latent variable model for sequential data.
\newblock In Cortes, C., Lawrence, N.~D., Lee, D.~D., Sugiyama, M., and
  Garnett, R., editors, {\em Advances in Neural Information Processing Systems
  28}, pages 2980--2988. Curran Associates, Inc.

\bibitem[Goodfellow et~al., 2014]{goodfellow2014generative}
Goodfellow, I., Pouget-Abadie, J., Mirza, M., Xu, B., Warde-Farley, D., Ozair,
  S., Courville, A., and Bengio, Y. (2014).
\newblock Generative adversarial nets.
\newblock In {\em Advances in neural information processing systems}, pages
  2672--2680.

\bibitem[Goodfellow et~al., 2013]{MaxoutNetworks_2013}
Goodfellow, I.~J., Warde-Farley, D., Mirza, M., Courville, A., and Bengio, Y.
  (2013).
\newblock Maxout networks.
\newblock In {\em Proceedings of the 30th International Conference on
  International Conference on Machine Learning}, pages III--1319--III--1327.

\bibitem[Hinton, 2002]{Hinton_Prod_of_Experts_Contrastive_2002}
Hinton, G.~E. (2002).
\newblock Training products of experts by minimizing contrastive divergence.
\newblock {\em Neural Comput.}, 14(8):1771--1800.

\bibitem[Hjelm et~al., 2018]{devon2018boundary}
Hjelm, R.~D., Jacob, A.~P., Trischler, A., Che, G., Cho, K., and Bengio, Y.
  (2018).
\newblock Boundary seeking {GAN}s.
\newblock In {\em International Conference on Learning Representations}.

\bibitem[Jozefowicz et~al., 2016]{limits_of_NLP_2016}
Jozefowicz, R., Vinyals, O., Schuster, M., Shazeer, N., and Wu, Y. (2016).
\newblock Exploring the limits of language modeling.
\newblock {\em CoRR}, abs/1602.02410.

\bibitem[Karras et~al., 2018]{karras2018progressive}
Karras, T., Aila, T., Laine, S., and Lehtinen, J. (2018).
\newblock Progressive growing of {GAN}s for improved quality, stability, and
  variation.
\newblock In {\em International Conference on Learning Representations}.

\bibitem[Kingma and Welling, 2014]{VAE_generative_model_14}
Kingma, D.~P. and Welling, M. (2014).
\newblock Auto-encoding variational bayes.
\newblock {\em Proceeding of International Conference on Learning
  Representations 2014}.

\bibitem[Rolfe, 2017]{discrete_VAE_2017}
Rolfe, J.~T. (2017).
\newblock Discrete variational autoencoders.
\newblock In {\em ICLR}.

\bibitem[Salakhutdinov and Hinton, 2009]{pmlr-v5-salakhutdinov09a}
Salakhutdinov, R. and Hinton, G. (2009).
\newblock Deep boltzmann machines.
\newblock In van Dyk, D. and Welling, M., editors, {\em Proceedings of the
  Twelth International Conference on Artificial Intelligence and Statistics},
  volume~5 of {\em Proceedings of Machine Learning Research}, pages 448--455,
  Hilton Clearwater Beach Resort, Clearwater Beach, Florida USA. PMLR.

\bibitem[Salimans et~al., 2016]{salimans2016improved}
Salimans, T., Goodfellow, I., Zaremba, W., Cheung, V., Radford, A., and Chen,
  X. (2016).
\newblock Improved techniques for training gans.
\newblock In {\em Advances in Neural Information Processing Systems}, pages
  2234--2242.

\bibitem[Shazeer et~al., 2017]{sparsely_gated_moe_NN_2017}
Shazeer, N., Mirhoseini, A., Maziarz, K., Davis, A., Le, Q., Hinton, G., and
  Dean, J. (2017).
\newblock Outrageously large neural networks: The sparsely-gated
  mixture-of-experts layer.
\newblock In {\em ICLR}.

\bibitem[Srivastava et~al., 2015]{HighwayNetworks_2015}
Srivastava, R.~K., Greff, K., and Schmidhuber, J. (2015).
\newblock Highway networks.
\newblock {\em CoRR}, abs/1505.00387.

\bibitem[Tieleman, 2008]{Tieleman2008}
Tieleman, T. (2008).
\newblock Training restricted boltzmann machines using approximations to the
  likelihood gradient.
\newblock In {\em Proceedings of the 25th International Conference on Machine
  Learning}, pages 1064--1071, New York, NY, USA. ACM.

\bibitem[Vahdat et~al., 2018]{NIPS2018_Vahdat}
Vahdat, A., Andriyash, E., and Macready, W. (2018).
\newblock Dvae\#: Discrete variational autoencoders with relaxed boltzmann
  priors.
\newblock In Bengio, S., Wallach, H., Larochelle, H., Grauman, K.,
  Cesa-Bianchi, N., and Garnett, R., editors, {\em Advances in Neural
  Information Processing Systems 31}, pages 1864--1874. Curran Associates, Inc.

\bibitem[van~den Oord et~al., 2016]{Conditional_PixelCNN_NIPS2016}
van~den Oord, A., Kalchbrenner, N., Espeholt, L., kavukcuoglu, k., Vinyals, O.,
  and Graves, A. (2016).
\newblock Conditional image generation with pixel-cnn decoders.
\newblock In Lee, D.~D., Sugiyama, M., Luxburg, U.~V., Guyon, I., and Garnett,
  R., editors, {\em Advances in Neural Information Processing Systems 29},
  pages 4790--4798. Curran Associates, Inc.

\bibitem[van~den Oord et~al., 2017]{NIPS2017_vinyals}
van~den Oord, A., Vinyals, O., and kavukcuoglu, k. (2017).
\newblock Neural discrete representation learning.
\newblock In Guyon, I., Luxburg, U.~V., Bengio, S., Wallach, H., Fergus, R.,
  Vishwanathan, S., and Garnett, R., editors, {\em Advances in Neural
  Information Processing Systems 30}, pages 6306--6315. Curran Associates, Inc.

\bibitem[Zilly et~al., 2017]{RecurrentHighwayNetworks_2017}
Zilly, J.~G., Srivastava, R.~K., Koutn{\'{\i}}k, J., and Schmidhuber, J.
  (2017).
\newblock Recurrent highway networks.
\newblock In {\em {ICML}}, volume~70 of {\em Proceedings of Machine Learning
  Research}, pages 4189--4198.

\end{thebibliography}
\clearpage
\newpage

\appendix
\pagebreak

\section{Approximate Gradient Calculation for the Two-Layer Network}
\label{sec:mcmc}

In this section, we propose an approximate method to compute the gradients of the empirical log-likelihood of the two-layer switch network with respect to the model parameters.

\subsection{Likelihood Gradient}
Consider the empirical log-likelihood $L^{(k)}$ from~\eqref{eq:twolayer_Lk}. Differentiating $L^{(k)}$ with respect to the parameters of the first layer, $\tk_1$, yields the gradient

{\footnotesize
\begin{equation}
  \label{eq:grad-Lk-tk1}
\begin{aligned}
  \frac{\partial \Lk}{\partial \tk_1} &= \frac{1}{I} \sum_{i=1}^I \sum_{\fk_{[1:l]}} \frac{p(x^{(i)}_{k+1} | \fk_{[1:l]}; \tk_2) \frac{\partial p(\fk_{[1:l]} | x^{(i)}_{[1:k]}; \tk_1)}{\partial \tk_1} }{p(x^{(i)}_{k+1} | x^{(i)}_{[1:k]}; \tk)} \\
                                      &= \frac{1}{I} \sum_{i=1}^I \sum_{\fk_{[1:l]}} \frac{p(x^{(i)}_{k+1} | \fk_{[1:l]}; \tk_2)  p(\fk_{[1:l]} | x^{(i)}_{[1:k]}; \tk_1)}{p(x^{(i)}_{k+1} | x^{(i)}_{[1:k]}; \tk)} \\
  &\qquad \qquad \times \frac{\partial}{\partial \tk_1} \log p(\fk_{[1:l]} | x^{(i)}_{[1:k]}; \tk_1).
\end{aligned}
\end{equation}
}

Note that for each $1 \leq i \leq I$, the inner summation over $\fk_{[1:l]}$
could be written as an expectation with respect to the distribution $p_i$ on
$\Fk_{[1:l]}$ defined as
\begin{equation*}
\begin{aligned}
 p_i(\fk_{[1:l]}; \tk):=   \frac{p(x^{(i)}_{k+1} | \fk_{[1:l]}; \tk_2)  p(\fk_{[1:l]} | x^{(i)}_{[1:k]}; \tk_1)}{W_i(\tk)},
\end{aligned}
\end{equation*}
with the normalizing constant
\begin{equation*}
\begin{aligned}
  W_i(\tk) := \sum_{\tfk_{[1:l]}} p(x^{(i)}_{k+1} | \tfk_{[1:l]}; \tk_2)  p(\tfk_{[1:l]} | x^{(i)}_{[1:k]}; \tk_1).
\end{aligned}
\end{equation*}
Therefore, we may rewrite~\eqref{eq:grad-Lk-tk1} as
\begin{equation}
\label{eq:Lk-grad-theta1-exp}
\begin{aligned}
\!\!\!   \frac{\partial \Lk}{\partial \tk_1} = \frac{1}{I} \sum_{i=1}^I \evwrt{p_i}{\frac{\partial}{\partial \tk_1} \log p(\Fk_{[1:l]} | x^{(i)}_{[1:k]}; \tk_1)},
\end{aligned}
\vspace{0.1in}
\end{equation}
where in the expectation on the RHS, $\Fk_{[1:l]}$ has the distribution $p_i$ defined
above. A similar calculation shows that
\begin{equation}
  \label{eq:Lk-grad-theta2-exp}
\begin{aligned}
\!\!\!  \frac{\partial \Lk}{\partial \tk_2} = \frac{1}{I} \sum_{i=1}^I \evwrt{p_i}{\frac{\partial}{\partial \tk_2} \log p(x^{(i)}_{k+1} | \Fk_{[1:l]} ; \tk_2)}.
\end{aligned}
\vspace{0.1in}
\end{equation}

\begin{figure}[t]
  \centering
  \input{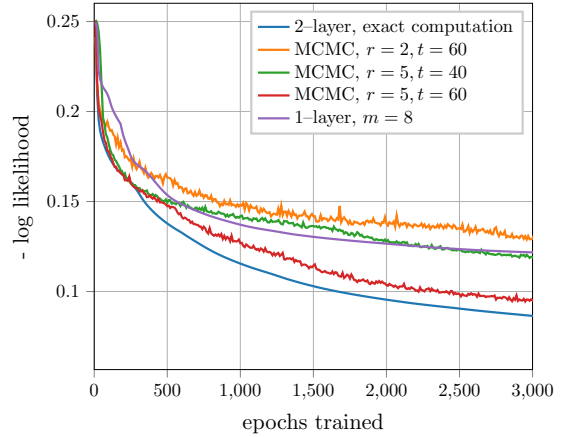}
  \caption{The performance of the MCMC algorithm for the two-layer switch
    network for bit index $629$ of the \texttt{MNIST} data. The size of the  switch
    network in this example is $(m_1, l, m_2) = (8, 8, 4)$. We run the Markov
    chain $t$ steps for $r$ independent runs and take the average to approximate
  the gradients. By increasing the values of $r$ and $t$, the likelihood performance approaches that of the exact gradient computation. The performance of the single-layer switch network with $m = 8$ is illustrated for reference. }
  \label{fig:mcmc}
\end{figure}

\subsection{Metropolis-Hastings Algorithm}
The Metropolis--Hastings algorithm may be used to approximate the expectation for each $i$. To see this, note that the ratio of probabilities
\begin{equation*}
\begin{aligned}
\frac{p_i(\fk_{[1:l]}; \tk)} { p_i(\tfk_{[1:l]}; \tk) }
\end{aligned}
\end{equation*} 
for two configurations $\fk_{[1:l]}$ and $\tfk_{[1:l]}$ does not depend on the normalizing constant $W_i(\tk)$ and can be computed efficiently. More precisely, for $1 \leq i \leq I$, we design a Markov Chain Monte Carlo (MCMC) algorithm with the proposal distribution
\begin{equation*}
\vspace{0.1in}
\begin{aligned}
  g_i(\tfk_{[1:l]} | \fk_{[1:l]}) = p(\tfk_{[1:l]} | x^{(i)}_{[1:k]}; \tk_1).
\end{aligned}
\vspace{0.05in}
\end{equation*}
With the current sample $\fk_{[1:l]}$ and the new sample $\tfk_{[1:l]}$, the
acceptance ratio takes the following form
\vspace{0.1in}
\begin{align*}
  A_i(\tfk_{1:l} | \fk_{[1:l]}) &= \min \left ( 1, \frac{p_i(\tfk_{[1:l]}; \tk)}{p_i(\fk_{[1:l]}; \tk)} \frac{g_i(\fk_{[1:l]} | \tfk_{[1:l]})}{g_i(\tfk_{[1:l]} | \fk_{[1:l]})} \right ) \\
  &= \min \left( 1, \frac{p(x^{(i)}_{k+1} | \tfk_{[1:l]}; \tk_2)}{p(x^{(i)}_{k+1} | \fk_{[1:l]}; \tk_2)} \right).
\end{align*}
\subsubsubsection{\bf{Discussion:}} We can interpret this procedure as generating samples given the previous $k$ symbols, and then re--weighting based on the likelihood of the symbol $k+1$. With this setup, we iterate the above Markov chain for $t$ steps, and repeat this procedure for $r$ rounds. Finally, we take the average of these independent $r$ outcomes, where each of them is the result of a Markov chain after $t$ iterations, to approximate each term in the gradients of~\eqref{eq:Lk-grad-theta1-exp} and \eqref{eq:Lk-grad-theta2-exp}. Figure~\ref{fig:mcmc} illustrates the performance of this approach for one bit in the \texttt{MNIST} dataset and for different values of the parameters $r$ and $t$. As we can see, by increasing the values of $r$ and $t$, the approximate gradient converges to the actual gradient.


\end{document}